\documentclass[11pt,letterpaper]{article}
\usepackage{naaclhlt2016}
\usepackage{times}
\usepackage{latexsym}
\usepackage{graphicx}
\usepackage{float}
\usepackage{url}

\naaclfinalcopy % Uncomment this line for the final submission
\def\naaclpaperid{***} %  Enter the naacl Paper ID here

\title{MITRE at SemEval-2016 Task 6: Transfer Learning for Stance Detection }

\author{Guido Zarrella \and Amy Marsh\\
	    The MITRE Corporation\\
	    202 Burlington Road\\
	    Bedford, MA 01730-1420, USA\\
	    {\tt {jzarrella,amarsh}@mitre.org}}

\date{}

\begin{document}

\maketitle

\begin{abstract}
We describe MITRE's submission to the SemEval-2016 Task 6, Detecting Stance in Tweets. This effort achieved the top score in Task A on supervised stance detection, producing an average F1 score of 67.8 when assessing whether a tweet author was in favor or against a topic. We employed a recurrent neural network initialized with features learned via distant supervision on two large unlabeled datasets. We trained embeddings of words and phrases with the word2vec skip-gram method, then used those features to learn sentence representations via a hashtag prediction auxiliary task. These sentence vectors were then fine-tuned for stance detection on several hundred labeled examples. The result was a high performing system that used transfer learning to maximize the value of the available training data. \end{abstract}

\section{Introduction}

This paper describes a system for performing automatic stance detection in social media messages. Our approach employs a recurrent neural network which was initialized from pre-trained features learned in successive attempts to encode world knowledge via weak external supervision.

Stance detection is the task of determining whether the author of a text is in favor or against a given topic, while rejecting texts in which neither inference is likely. This task is distinct from sentiment analysis in that an \textit{in favor} or \textit{against} stance can be measured independently of an author's emotional state. In stance detection we attempt to measure how an author's opinion is expressed in spontaneous, unstructured messages rather than the explicit prompts of formal opinion polls.

Declarations of stance are often couched in figurative language that can be difficult for machines to unravel. Consider the texts \textit{We don't inherit the earth from our parents we borrow it from our children} and \textit{Last time I checked, Al Gore is a politician, not a scientist}. 
To the human observer messages like these contain an interpretable stance relevant to the topic of climate change.
But to understand rhetorical devices like sarcasm, irony, analogy, and metaphor, a reader often uses personal experience to infer broader context.
For machines, matters are additionally complicated by use of informal vocabulary, grammar, and spelling. 
Furthermore, training data is often expensive or difficult to collect in bulk. 
These challenges motivated our efforts to seek transfer learning of broad world knowledge through feature pre-training using large unlabeled datasets.

\section{Related Work}

It is common for machine learning approaches to begin learning of any new task from scratch, for example by randomly initializing the parameters of a neural network. This disregards any knowledge gained by similar algorithms when solving previous tasks. Transfer learning approaches, on the other hand, store the knowledge gained in one context and apply it to different, related problems. This type of approach is particularly appealing when one lacks sufficient quantity of in-domain labeled training data, such as when there are only a few hundred known examples of a target. 

One strategy for performing transfer learning is to train the parameters of a neural network on multiple tasks: first on an auxiliary task with plentiful data that allows the network to identify meaningful features present in the corpus, then a second time using actual task data to tune and exploit those features learned in the first pass.

Deep neural networks trained for image classification can be improved when initialized with features learned from distant tasks, for example Yosinski et al. ~\shortcite{Yosinski:14}. In natural language processing domains, sentence representations learned on unlabeled data have been shown to be useful across a variety of classification and semantic similarity tasks ~\cite{kiros:15,dai:15,hill:16}. Weston et al.~\shortcite{Weston:14} used a hashtag prediction task to learn sentence representations that improve a downstream content-based recommendation system. 

Previous work in stance detection is significant \cite{Mohammad:15}, often with a focus on analysis of congressional debates or online forums \cite{thomas:2006,somasundaran:2009,murakami:2010,walker:2012} in which discourse and dialogue features offer clues for identifying oppositional speakers. Rajadesingan and Liu \shortcite{rajadesingan:2014} study stance detection in Twitter conversations and use a retweet-based label propagation approach. This objective of this work differs in that we attempt to detect an author's stance purely from analysis of the text of a single message.

\section{Task and Evaluation}
\textit{Detecting Stance in Tweets, Subtask A: Supervised Frameworks} \cite{StanceSemEval2016} was a shared task organized within SemEval-2016.

The task organizers provided training data in the form of 2,814 tweets covering five topics, with 395 to 664 tweets per topic. The organizers used crowdsourcing to manually annotate these tweets for stance. Class balance varied between topics, with some topics showing significant skew (e.g. \textit{Climate Change is a Real Concern} with 4\% AGAINST and 54\% FAVOR) while others were more balanced (e.g. \textit{Feminist Movement} with 49\% AGAINST and 32\% FAVOR). Approximately 74\% of the provided tweets were judged to be either in favor or against, while the remainder contained neither inference. An additional 1249 tweets with held-out labels were used as evaluation data. Systems were evaluated using the macro-average of F1-score(FAVOR) and F1-score(AGAINST) across all topics.

\section{System Overview}
\label{systemoverview}
\begin{figure}
  \includegraphics[width=\linewidth]{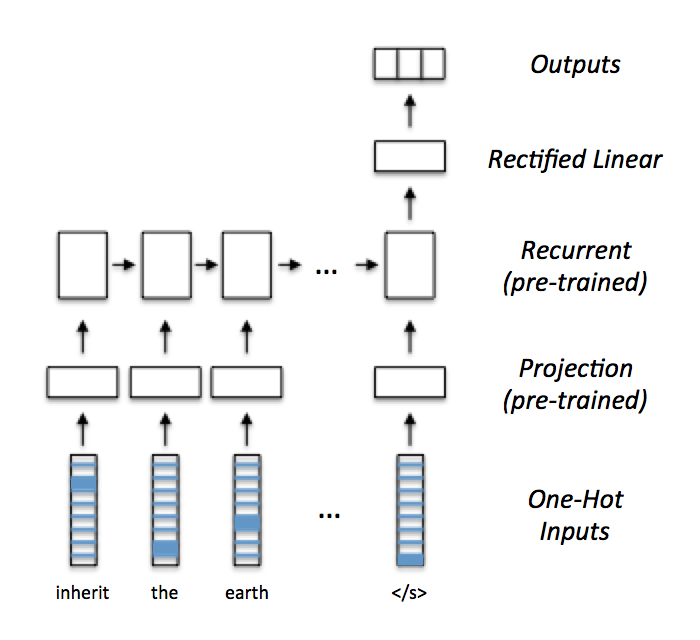}
  \caption{A recurrent neural network for stance detection.}
  \label{tab:arch}
\end{figure}

We now describe an approach to stance detection that employs a recurrent neural network organized into four layers of weights (shown in Figure \ref{tab:arch}). Input tokens are encoded in a one-hot fashion, such that each token is represented by a sparse binary vector containing a single one-value at the index corresponding to the token's position in the vocabulary. A sequence of these inputs are projected through a 256-dimensional embedding layer, which feeds into a recurrent layer containing 128 Long Short-Term Memory (LSTM) units. The terminal output of this recurrent layer is densely connected to a 128-dimensional layer of Rectified Linear units trained with 90\% dropout ~\cite{srivastava:2014}. Finally, this layer is fully connected to a three dimensional softmax layer in which each unit represents one of the output classes: \textit{FAVOR}, \textit{AGAINST}, or \textit{NONE}. 

This approach did not incorporate any manually engineered task-specific features or inputs relevant to the surface structure of the text. The only inputs to the network were the sequence of indices representing the identity of lowercased tokens (words or phrases) in the text. All feature pre-training was done using weak supervision from larger unlabeled text datasets, with a goal of automatically learning useful representations of words and input sequences. 

\subsection{Pre-Training the Projection Layer}
The weights for the projection layer of the network were initialized from 256-dimensional word embeddings learned using the \texttt{word2vec} skip-gram \cite{Mikolov:13a} algorithm.
We sampled 218,179,858 tweets from Twitter’s public streaming API during 2015, and used this unlabeled data as our training corpus. 
Retweets, duplicates, and non-English messages were not included in this sample.
Text was lowercased and tokenized to mimic the style of the task data.
We then applied \texttt{word2phrase} \cite{Mikolov:13b} twice consecutively to identify phrases comprised of up to four words, for example making a single token of the phrase \textit{global climate change}. 

We then trained 256-dimensional skip-gram embeddings for the 537,366 vocabulary items that appeared at least 100 times in our corpus, with a context window of 10 words and 15 negative samples per positive example.  
These hyperparameters were chosen in advance based on our prior experience in training embeddings for identifying word analogies and estimating semantic similarity of sentences.
Out of vocabulary items were represented by the average of all in-vocabulary vectors.

Note that these projection layer weights were later tuned by backpropagation during training of the recurrent networks. 
Thus these initializations served to provide the RNNs with initial feature representations intended to capture the nuances of informal word usage observed in a large sample of text.

\begin{figure*}
  \includegraphics[width=\textwidth]{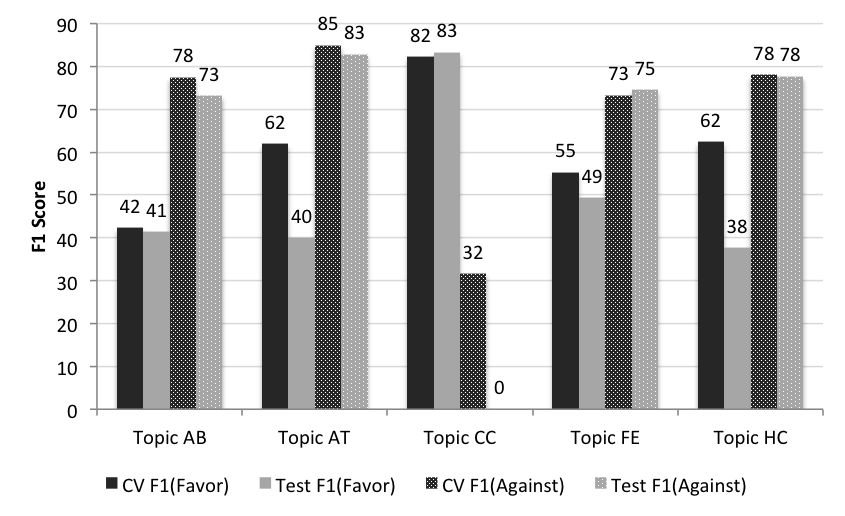}
  \caption{F1 scores for each topic and class on both cross-validation and test conditions.}
\label{fig:breakdown}
\end{figure*}

\subsection{Pre-Training the Recurrent Layer}
The second layer of our network was composed of 128 Long Short-Term Memory (LSTM) units \cite{Hochreiter:1997}.
This recurrent layer received as input a sequence of up to 30 embeddings, folding each into its hidden state in turn.
It was initialized with weights that were pre-trained using the distant supervision of a hashtag prediction auxiliary task.
In this manner the network learned distributed sentence representations from a dataset containing a broad array of stance declarations, rather than relying exclusively on the 2,814 explicitly labeled in-domain tweets.

We began by automatically identifying 197 hashtags with relevance to the topics under consideration, for example \textit{\#climatechange}, \textit{\#climatescam}, and \textit{\#gamergate}.
These hashtags were selected on the basis of a nearest-neighbor search of the word embedding space.
We queried the vector space using the embeddings of the topic titles, and selected the unique hashtags with high (top-50) cosine similarity.
These selections varied greatly in frequency and task specificity, including a number of tags which were related to multiple topics and others which appeared ambiguously related. 
Half of the 40 most frequent tags in this list were related to the 2016 United States presidential elections.
The final list of 197 relevant hashtags was held constant across all experiments.

We extracted 298,973 tweets containing at least one of these 197 hashtags from the 2015 corpus of 218 million English tweets. 
Text was lowercased, tokenized, and phrase chunked according with the preprocessing choices made during the training of word embeddings. 
If a tweet contained more than one hashtag, the most frequent tag was used as the prediction target. 
Tweets were then stripped of all hashtags, including both the correct hashtag and any additional hashtags appearing in the tweet.  

This corpus was divided into a training set and development set using a 90/10 split.  Each word in the tweet was converted into a vector using the word embeddings.  The sequence of vector representations of the words in the tweet served as the input to a neural network with a 128-dimensional LSTM layer, followed by a dense softmax layer over the 197 possible candidate hashtags.  

We trained the neural network with gradient descent using AdaDelta and categorical cross entropy minimization. Both the word embeddings and the recurrent layer were tuned during this process. Training continued until the accuracy on the development set reached its maximum, which took seven epochs. The final model correctly predicted development set hashtags with 42.6\% accuracy.

\section{Experiments}
The system described in section~\ref{systemoverview} was designed to detect stances pertaining to a single topic. 
As such we trained five distinct classifiers, one for each of the five topics under consideration in the evaluation.
The embedding and recurrent layers of each classifier were initialized with the weights obtained from the pre-training process described above. 
The remainder of the weights were randomly initialized and the network was trained with stochastic gradient descent using a learning rate of 0.015 and momentum of 0.9.
These networks were trained using a categorical cross-entropy loss function, with costs for each example weighted according to the prevalence of the class in the training data. This placed higher weight on rare classes.  
The recurrent networks were implemented using the \texttt{Keras} framework \cite{chollet:2015}.

The training data for each topic was shuffled and split into five chunks for cross-validation. 
The training process for a single topic's classifier therefore resulted in five distinct neural networks, each learning from 80\% of the training data. 
These training set sizes ranged from 316 to 532 tweets. 
Each network was trained for 50 epochs, with early stopping to select the model with the best validation loss. 
Predictions from these five trained networks were used to select a single class via majority vote at decode time. 

Variants of this approach were considered as well. One variant used an identical framework with a recurrent layer initialized instead from a RNN trained on 6.5 million tweets containing the top 10,000 most frequent hashtags (as opposed to 197 topic-relevant hashtags). We also omitted the RNN pre-training altogether and randomly initialized the recurrent layer. These variants were not found to improve performance.

\section{Results}
Our submission achieved an average F1 score of 67.8 on the \textit{FAVOR} and \textit{AGAINST} classes of the held out test set, which contained tweets from all five topics. This was the top scoring system among the 19 entries submitted to the supervised stance detection shared task. 

This same system had an average F1 of 71.1 in testing of the component systems using cross-validation on the training set, indicating a small amount of overfitting. Scores also varied moderately across topics and classes (Figure ~\ref{fig:breakdown}).

One consistent observation across all topics was that the majority class, whether it was \textit{FAVOR} or \textit{AGAINST}, significantly outperformed the corresponding minority class. There was positive correlation ($R^2$ = 0.67) between the F1 score for a given class and the raw number of training examples representing that class. 

The weight pre-training and initialization regimes that we applied improved performance relative to the tested alternatives. Entirely omitting pre-training of the recurrent layer (while keeping the projection layer pre-training) resulted in a drop of average F1 from 71.1 to 70.0 in 5-fold cross-validation. Meanwhile the RNN trained to select from among 10,000 popular hashtags led to an average F1 of 66.0, a relative reduction of 7.2\%  compared to the submission initialized from the RNN trained on 197 highly relevant hashtags.  

\section{Conclusion}
We described a state-of-the-art system for automatically determining the stance of an author based on the content of a single tweet. This approach was able to maximize the value of limited training data by transferring features from other systems trained on large, unlabeled datasets.

Our results demonstrated that hashtag prediction and skip-gram tasks can result in pre-trained features that are useful for stance detection. The selection of domain-relevant hashtags appears to be a crucial aspect of this architecture, as experiments employing a larger collection of frequent hashtags resulted in significantly worse performance on the stance detection task. 

Transfer learning does not completely eliminate the need for labeled in-domain training data. The most frequent stance classes uniformly outperformed the minority classes by all metrics. It is likely that stances which are rare in this training set are also proportionally absent from the larger unlabeled auxiliary hashtag task. Future experiments could investigate other techniques for identifying relevant hashtags, with a goal of maximizing the diversity of opinions represented in the auxiliary datasets.

\section*{Acknowledgments}
This work was funded under the MITRE Innovation Program. 
Thanks to Spencer Marsh for his timely encouragement. 
Approved for Public Release; Distribution Unlimited: Case Number 16-1159.
\copyright 2016 The MITRE Corporation: ALL RIGHTS RESERVED.
\bibliography{naaclhlt2016}
\bibliographystyle{naaclhlt2016}
\end{document}